\documentclass{article}

\PassOptionsToPackage{numbers, compress, sort}{natbib}

\usepackage[final]{neurips_2023}

\usepackage[utf8]{inputenc} 
\usepackage[T1]{fontenc}    
\usepackage{xcolor} 
\usepackage{hyperref}  
\usepackage{graphicx}
\hypersetup{
    colorlinks=true,
    citecolor=gray,
    }
\usepackage{url}            
\usepackage{booktabs}       
\usepackage{amsmath,amssymb}       
\usepackage{nicefrac}       
\usepackage{microtype}      
\usepackage[all=normal, mathspacing=tight,mathdisplays=normal, floats=tight, paragraphs=tight, lists=normal]{savetrees}
\renewcommand{\b}{\boldsymbol}
\usepackage[capitalize]{cleveref}
\crefname{section}{Sec.}{Secs.}
\Crefname{section}{Section}{Sections}
\Crefname{table}{Table}{Tables}
\crefname{table}{Tab.}{Tabs.}
\crefname{algorithm}{Alg.}{Algs.}
\title{Score Normalization for a \\ Faster Diffusion Exponential Integrator Sampler}

\author{%
Guoxuan Xia$^1$\thanks{Work done whilst interning at MediaTek Research.} \quad Duolikun Danier$^{2*}$\quad Ayan Das$^3$\quad Stathi Fotiadis$^{1,3}$\\ \textbf{Farhang Nabiei$^3$\quad Ushnish Sengupta$^3$\quad Alberto Bernacchia$^3$}\\
$^1$Imperial College London, UK \quad
$^2$University of Bristol, UK \quad
$^3$MediaTek Research, UK \\
\texttt{g.xia21@imperial.ac.uk}\quad \texttt{duolikun.danier@bristol.ac.uk}\\\texttt{\{first.last\}@mtkresearch.com}
}

\begin{document}

\maketitle

\begin{abstract}


Recently, \citet{zhang2023fast} have proposed the Diffusion Exponential Integrator Sampler (DEIS) for fast generation of samples from Diffusion Models. It leverages the \textit{semi-linear} nature of the probability flow ordinary differential equation (ODE) in order to greatly reduce integration error and improve generation quality at low numbers of function evaluations (NFEs). Key to this approach is the \textit{score function reparameterisation}, which reduces the integration error incurred from using a fixed score function estimate over each integration step. The original authors use the default parameterisation used by models trained for noise prediction -- multiply the score by the standard deviation of the conditional forward noising distribution. We find that although the mean absolute value of this score parameterisation is close to constant for a large portion of the reverse sampling process, it changes rapidly at the end of sampling. As a simple fix, we propose to instead reparameterise the score (at inference) by dividing it by the average absolute value of previous score estimates at that time step collected from offline high NFE generations. We find that our score normalisation (DEIS-SN) consistently improves FID compared to vanilla DEIS, showing an improvement at 10 NFEs from 6.44 to 5.57 on CIFAR-10 and from 5.9 to 4.95 on LSUN-Church ($64\times 64$). Our code is available at \url{https://github.com/mtkresearch/Diffusion-DEIS-SN}.
\end{abstract}

\section{Introduction}
Diffusion models \cite{ddpm, sohl} have emerged as a powerful class of deep generative models, due to their ability to generate diverse, high-quality samples, rivaling the performance of GANs \cite{stylegan2} and autoregressive models \cite{PixeCNN}. They have shown promising results across a wide variety of domains and applications including (but not limited to) image generation \cite{dhariwal2021diffusion, rombach2022high}, audio synthesis \cite{liu2023audioldm}, molecular graph generation \cite{hoogeboom2022equivariant}, and 3D shape generation \cite{poole2022dreamfusion}. They work by gradually adding Gaussian noise to data through a forward diffusion process parameterized by a Markov chain, and then reversing this process via a learned reverse diffusion model to produce high-quality samples. 

The sampling process in diffusion models can be computationally expensive, as it typically requires hundreds to thousands of neural network function evaluations to generate high-quality results. Moreover, these network evaluations are necessarily \textit{sequential}, seriously hampering generation latency. This has motivated recent research efforts to develop acceleration approaches that allow the generation of samples with fewer function evaluations while maintaining high sample quality. 
Denoising Diffusion Implicit Models (DDIM) \cite{ddim} is an early but effective acceleration approach that proposes a non-Markovian noising process for more efficient sampling. \citet{score_based} show that samples can be quickly generated by using black-box solvers to integrate a probability flow ordinary differential equation (ODE). 
Differentiable Diffusion Sampler Search (DDSS) \cite{watson2021learning} treats the design of fast samplers for diffusion models as a differentiable optimization problem. 
 Diffusion model sampling with neural operator (DSNO) \cite{diffusion_operator} accelerates the sampling process of diffusion models by using neural operators to solve probability flow differential equations. Progressive Distillation \cite{progdist} introduces a method to accelerate the sampling process by iteratively distilling the knowledge of the original diffusion model into a series of models that learn to cover progressively larger and larger time step sizes.

Recently, a new family of fast samplers, that leverage the \textit{semi-linear} nature of the probability flow ODE, have enabled new state-of-the-art results at low numbers of function evaluations (NFEs) \cite{dpm_solver,zhang2023gddim,zhang2023fast}. In this work we focus on improving the Diffusion Exponential Integrator Sampler (DEIS) \cite{zhang2023fast}, specifically the time-based Adams-Bashforth version (DEIS-$t$AB). Our contributions are as follows:
\begin{enumerate}
    \item We show that DEIS's default score parameterisation's average absolute value varies rapidly near the end of the reverse process, potentially leading to additional integration error.
    \item We propose a simple new score parameterisation -- to normalise the score estimate using its average empirical absolute value at each timestep (computed from high NFE offline generations). This leads to \textit{consistent improvements in FID compared to vanilla DEIS}. 
\end{enumerate}
\begin{figure}
    \centering
    \includegraphics[width=\linewidth]{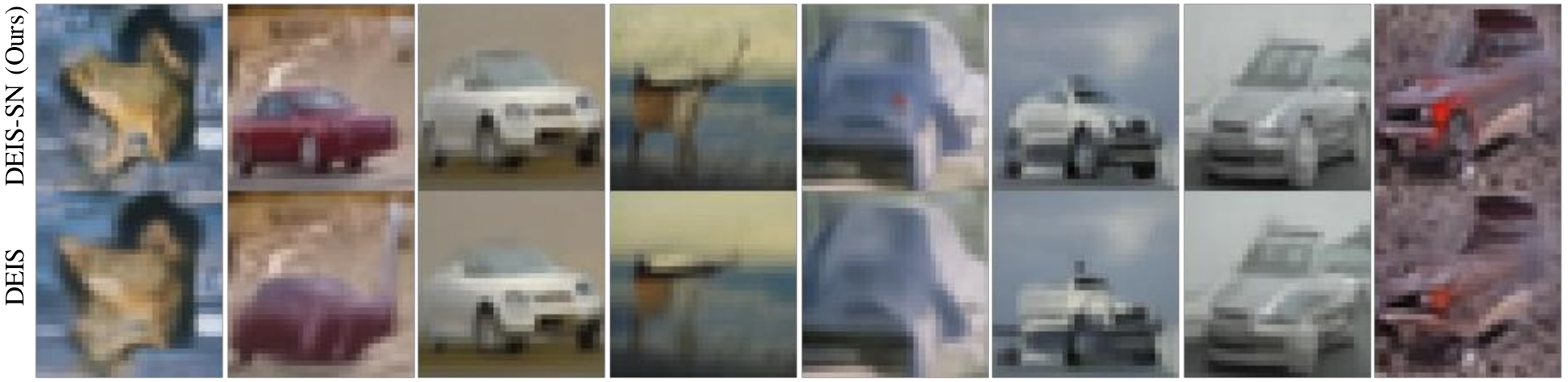}
    \caption{Generations at 5 NFEs with a CIFAR-10 model. \textcolor{gray}{Top}: DEIS-SN (ours). \textcolor{gray}{Bottom}: DEIS \cite{zhang2023fast}. It can be seen that DEIS-SN is often able to better generate details (such as wheels on cars). More generation results are shown in \cref{fig:unconditional_samples} of Appendix~\ref{app:more_gen_results}.}
    \label{fig:generation_comparison}
\end{figure}

\section{Preliminaries}
\paragraph{Forward Process} We define a \textit{forward process} over time $t\in [0,1]$ for random variable $\b x_t\in \mathbb R^D$,
\begin{equation}\label{eq:forward}
    p(\b x_t|\b x_0) = \mathcal{N}(\b x_t;\ a_t \b x_0,\; \sigma_t^2\b I)
\end{equation}
for $\b x_0$ drawn from some unknown distribution $p(\b x_0)$. Here, $a_t$ and $\sigma_t$ define the \textit{noise schedule}.\footnote{Note that we restrict ourselves to the case of isotropic noising as it applies to the vast majority of cases, although it is easy to generalise \cref{eq:forward} by replacing $a_t, \sigma^2_t$ with matrices $\b A_t, \b \Sigma_t$.} The following stochastic differential equation has the same conditional distributions as \cref{eq:forward} \cite{kingma2021on},
\begin{equation}\label{eq:sde}
    d\b x_t = f_t\b x_t dt + g_t d\b w_t, \quad \b x_0 \sim p(\b x_0),
\end{equation}
where $\b w_t \in \mathbb R^D$ is the Wiener process and
\begin{equation}
    f_t = \frac{d a_t}{dt}\frac{1}{a_t}, \quad g_t^2 = \frac{d \sigma_t^2}{dt} - 2f_t\sigma_t^2.
\end{equation}
\paragraph{Probability Flow ODE} \citet{score_based} show that the following ODE,
\begin{equation}\label{eq:ode}
    \frac{d \b x_t}{dt} = f_t \b x_t - \frac{1}{2}g^2_t\nabla_{\b x_t}\log\; p(\b x_t),  \quad \b x_1 \sim p(\b x_1),
\end{equation}
shares the same marginal distributions $p(\b x_t)$ as the stochastic differential equation in \cref{eq:sde}. Given a neural network that is trained to approximate the score $\b s_\theta(\b x_t, t) \approx \nabla_{\b x_t}\log\; p(\b x_t)$, \cref{eq:ode} can be exploited to generate samples approximately from $p(\b x_0)$ using blackbox ODE solvers \cite{score_based}. This approach tends to produce higher quality samples at lower NFEs vs stochastic samplers \cite{score_based}. We note that in the domain of diffusion models, neural networks tend to be trained to predict a \textit{parameterisation} of the score, e.g. $ \b s_\theta(\b x_t, t) = -\b \epsilon_\theta(\b x_t, t)/\sigma_t$ (noise prediction) \cite{ddpm} or $s_\theta(\b x_t, t) = (a_t \b x_\theta(\b x_t, t)-\b x_t)/\sigma_t^2$ (sample prediction) \cite{progdist}, as this leads to better optimisation and  generation quality \cite{ddpm}.

\paragraph{Exponential Integrator} Recently, a number of different approaches have exploited the \textit{semi-linear} structure of \cref{eq:ode}, solving the linear part, $f_t\b x_t$, exactly \cite{dpm_solver, zhang2023fast, zhang2023gddim, lu2023dpmsolver}. This has lead to impressive generation quality at low NFEs ($\leq 20$). One such approach is the Diffusion Exponential Integrator Sampler (DEIS)\footnote{Note that we focus only on the time-based version of DEIS, $t$AB.} \cite{zhang2023fast} that uses the following iteration over time grid $\{t_i\}_{i=0}^N$ to generate samples:\footnote{Note that we simplify notation compared to \cite{zhang2023fast} by restricting ourselves to the case of isotropic noise.}
\begin{align}\label{eq:ei}
    \b x_{t_{i-1}} &= \Psi(t_{i-1},t_i)\b x_{t_i} \;+\; \sum_{j=0} ^r C_{ij}\underbrace{(-K_{t_{i+j}})\b s_\theta(\b x_{t_{i+j}},t_{i+j})}_{\text{score reparameterisation}}, \\
    C_{ij} &= \int_{t_i}^{t_{i-1}} \frac{1}{2}\Psi(t_{i-1},\tau)g_\tau^2K_\tau^{-1}\prod_{k\neq j}\left[\frac{\tau - t_{i+k}}{t_{i+j}-t_{i+k}}\right]d\tau, \nonumber
\end{align}
where $\Psi(t,s)$ satisfies $\frac{\partial\Psi(t,s)}{\partial t}=f_t\Psi(t,s),\Psi(s,s)=1$, $K_t$ is a function used to reparameterise the score estimate and $r$ is the order of the polynomial used to extrapolate said score reparameterisation. We note that $C_{ij}$ and $\Psi(t,s)$ can be straightforwardly calculated offline using numerical methods.

\section{The Importance of Score Reparameterisation}
\begin{figure}
    \centering
    \includegraphics[width=0.9\linewidth]{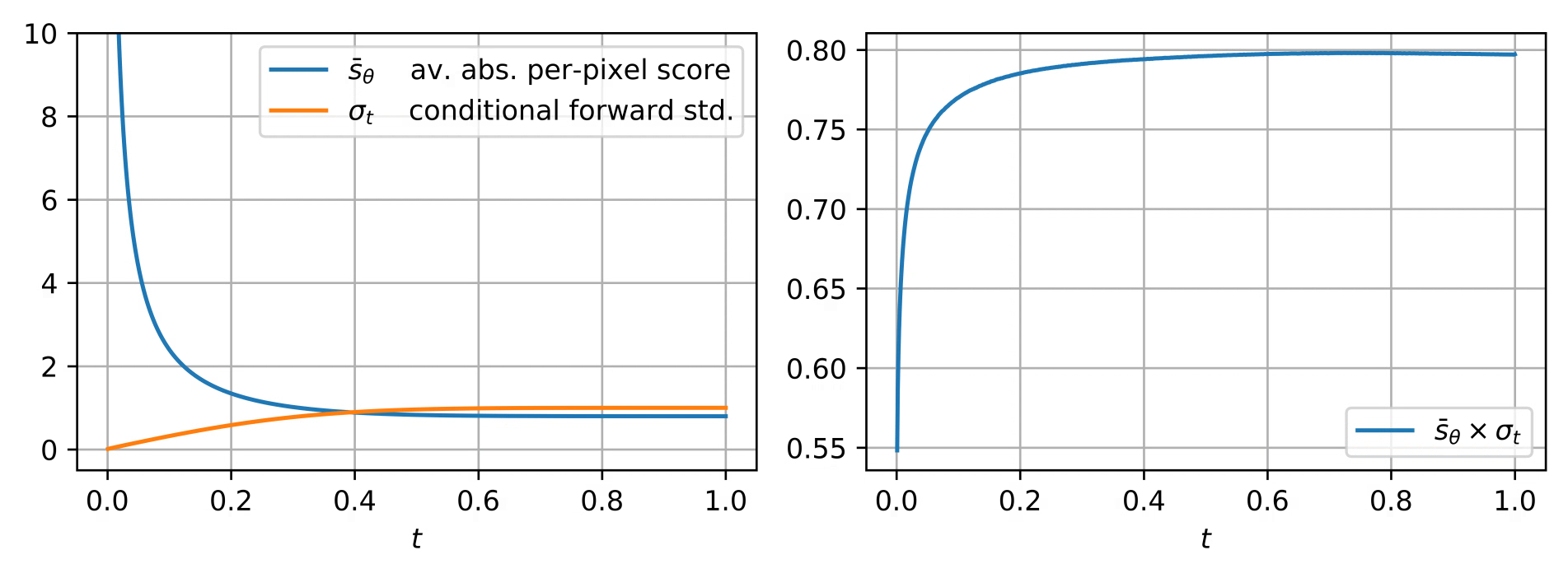}    
    \caption{\textcolor{gray}{Left}: average absolute per-pixel score estimate value $\bar{s}_{\theta}$ and the conditional forward standard deviation $\sigma_t$ plotted over the reverse process. \textcolor{gray}{Right}: the product of the previous two values -- we see that this is constant for most of the reverse process, but it changes rapidly near $t=0$. The noising schedule is the Linear $\beta$ schedule \cite{ddpm}, the training data is CIFAR-10 and we average 256 samples.}
    \label{fig:score_illust}
\end{figure}

The key to reducing integration error in \cref{eq:ei} is the score reparameterisation. DEIS approximates an integral over $\tau \in [t_{i-1}, t_i]$, over which $-K_{\tau}\b s_\theta(\b x_{\tau},\tau)$ should vary, using fixed estimates from finite $t$s. \textit{This approximation will be more accurate the less $-K_{\tau}\b s_\theta(\b x_{\tau},\tau)$ varies with $\tau$}. \citet{zhang2023fast} choose the default parameterisation for noise prediction models, $K_t=\sigma_t$, where $\b \epsilon_\theta(\b x_t, t)=-\sigma_t\b s_\theta(\b x_t, t)$. \cref{fig:score_illust} shows that this reparameterisation (right) is roughly constant in average absolute pixel value ($\bar{s}_\theta\sigma_t$) for the majority of the generation process,\footnote{This is corroborated in Fig. 4a of \cite{zhang2023fast}} whilst the score estimate varies considerably more. By reducing the variation over time of the score estimate via reparameterisation, DEIS is able to achieve much lower integration error and better quality generations at low NFEs \cite{zhang2023fast}.

\section{Score Normalisation (DEIS-SN)}
\cref{fig:score_illust} also shows, however, that near $t=0$ there is still substantial variation in $\bar{s}_\theta\sigma_t$. Thus we propose a new score reparameterisation, where we simply set $K_t=1/\bar{s}_\theta(t)$. That is to say, \textbf{at inference, we normalise the score estimate with the empirical average absolute pixel value at $t$ of previous score estimates.} The aim of this is to further reduce the variation in the reparameterisation near $t=0$.

$\bar{s}_\theta(t)$ can be found using offline generations at high NFEs. We use linear interpolation to accommodate continuous $t$. Our approach can be directly plugged into DEIS, so we refer to it as DEIS-SN.

We note that gDDIM \cite{zhang2023gddim} is another extension to DEIS, however this approach is specifically applicable to data distributions that are well modelled by a single Gaussian (such as the velocity of Critically-Damped Langevin Diffusion \cite{CLD}), which is not the case for image datasets like CIFAR. 

\section{Experimental Results}
\begin{figure}
    \centering
    \includegraphics[width=\linewidth]{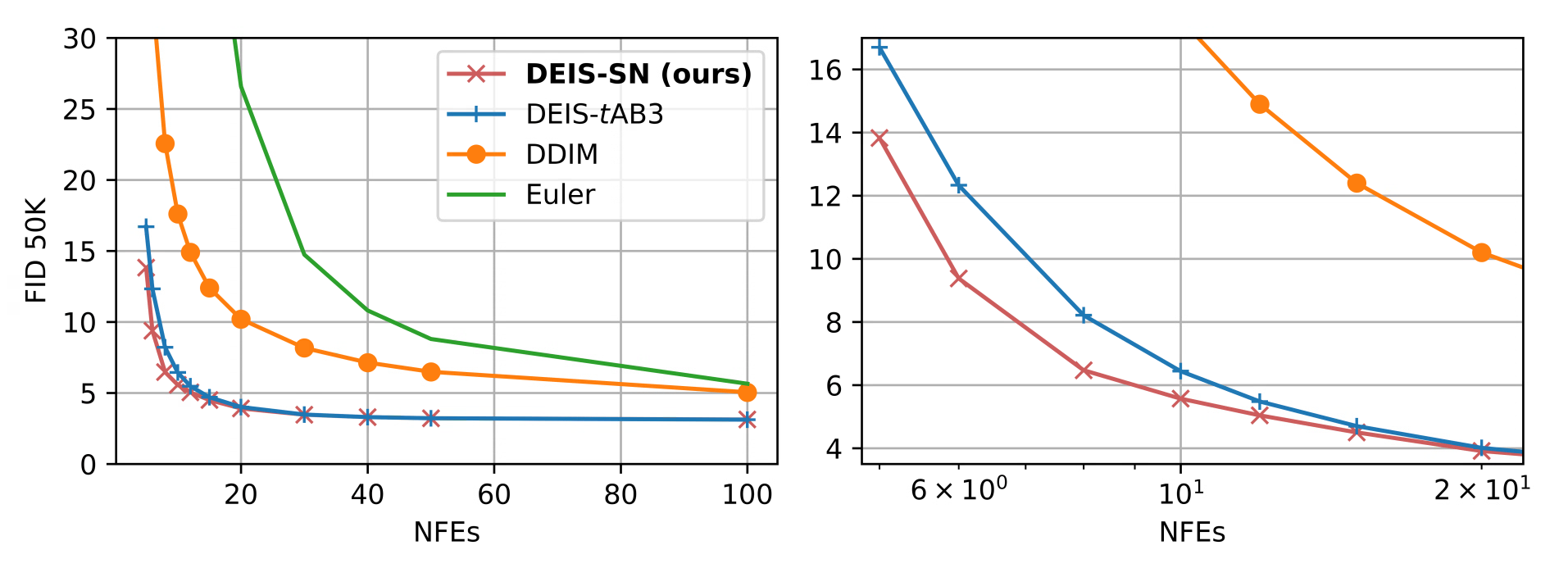}
    \caption{\textcolor{gray}{Left}: FID 50K 
 for CIFAR-10 against NFEs for different samplers on the same trained model. \textcolor{gray}{Right}: zoomed in view of the left figure. \textbf{DEIS-SN consistently outperforms DEIS at low NFEs.}}
    \label{fig:NFE-FID}
\end{figure}
 We train the UNet architecture from \cite{iddpm} on CIFAR-10 \cite{cifar} using the linear $\beta$ schedule \cite{ddpm} on noise prediction.\footnote{Note this is smaller than the architecture used in \cite{score_based,zhang2023fast} so the baseline high NFE FIDs are slightly worse.} We compare our approach to Euler integration of \cref{eq:ode}, DDIM \cite{ddim} and DEIS-$t$AB3 (polynomial order $r=3$) \cite{zhang2023fast}. DEIS-SN is identical to DEIS-$t$AB3 other than the choice of $K_t=1/\bar{s}_\theta(t)$ over $K_t=\sigma_t$. The average absolute score estimate $\bar{s}_\theta(t)$ is measured offline on a batch of 256 generations. For Euler and DDIM we use trailing linear time steps \cite{flawed}, whilst for DEIS we use trailing quadratic time $i\in \{0,\dots,N\}$, $t_i = (i/N)^2$. For full experimental details see \cref{app:exp}. A similar protocol was followed for LSUN-Church ($64\times 64$) experiments.

\cref{fig:NFE-FID} shows for CIFAR-10, that at low NFEs, DEIS-SN provides a consistent FID improvement over vanilla DEIS. For higher NFEs, when the width of the intervals $[t_{i-1}, t_i]$, and thus the integration error, is reduced, the benefit of DEIS-SN gradually disappears and DEIS-SN performs almost identically to vanilla DEIS. Both DDIM and Euler significantly underperform both DEIS approaches. Experimental results for LSUN-Church are shown in \cref{sec:lsunchurch}.

\cref{fig:generation_comparison} shows visual comparisons of vanilla DEIS and DEIS-SN at 5 NFEs.\footnote{We select examples with clear visual differences, as many generations are visually indistinguishable. (Nevertheless, the FID 50K results indicate there is a difference in generation quality on aggregate.)} We see that DEIS-SN can better generate details (such as vehicle wheels). We note that generally generations are visually similar on a high level. This is because the difference between vanilla DEIS and DEIS-SN only occurs for $t$ near 0 (\cref{fig:score_illust}), i.e. the end of the generation/reverse process.

\section{Conclusion}

In this work we propose to extend the Diffusion Exponential Integrator Sampler (DEIS) with empirical score normalisation (DEIS-SN). Through our novel score-reparameterisation, we aim to further reduce integration error towards the end of the generation process by normalising the score estimate with the empirical average absolute value of previous score estimates. We validate our approach empirically on CIFAR-10 and LSUN-Church, showing that DEIS-SN is able to consistently outperform vanilla DEIS for low NFE generations in terms of FID 50k. We also show visual examples of DEIS-SN's superiority. 

In the future it would be interesting to extend this work to cover non-isotropic cases, where the score reparameterisation $\b K_t$ is performed by a matrix, possibly in a transformed space such as the frequency domain \cite{spd,hoogeboom2023blurring}, to see if additional performance gains are to be had. Another possibility would be to parameterise $K_t$ and directly optimise it for better image quality at low NFEs.
\bibliographystyle{ieee_fullname_natbib}
\bibliography{bibliography}
\newpage
\appendix
\section{Additional Experimental Details}\label{app:exp}

In order to perform sampling with our proposed method, and for comparison purposes, we trained a model to approximate the true score $\nabla_{\b x_t} \log p(\b x_t)$ with a standard architecture and training procedure. We purposefully use a model with relatively small capacity to segregate the effects of sampling procedure and better score estimation \cite{score_based,elucidate}. We train our score estimator in a discrete DDPM \citet{ddpm} setup with time-discretization granularity $N=1000$, $i\in\{0,\dots,N\}$, $t_i=i/N$, which is considered to be a standard in diffusion literature. As suggested by \citet{ddpm}, we train the score estimator with the ``simple loss'' which is proven to be better at generation quality, as opposed to the true variational bound \cite{kingma2021on}. Again, as per \citet{ddpm}, we do not directly estimate the score $\b s_{\theta}$, but instead estimate the ``noise-predictor'' $\b \epsilon_{\theta}$. Concretely, We optimize the following objective,

\begin{equation} \label{eq:simple_loss_training}
    \mathbb{E}_{\b x_0\sim p(\b x_0),\ \b x_t\sim p(\b x_t |\b x_0),\ i\sim \mathcal{U}\{1, N\},\ \epsilon\sim\mathcal{N}(0, I)}\left[ || \b \epsilon_{\theta}(\b x_{t_i}, t_i) -\b \epsilon ||_2^2 \right],
\end{equation}

\noindent where $p(\b x_0)$ is the data distribution realized using our dataset, the forward noising conditional $p(\b x_t |\b  x_0)$ is from \cref{eq:forward} and $t_i$ are timesteps. We use the standard ``positional embeddings'' for incorporating $t_i$ into the noise-estimator neural network. The $(a_t, \sigma_t)$ in \cref{eq:forward} are chosen to be the standard ``linear schedule'' and \textit{variance-preserving} formulation \cite{score_based}, \emph{i.e.} 

\begin{equation}
    \textstyle a_t^2 = \prod_{t' = 1}^t (1 - \beta_{t'}),\text{ and } \sigma_t = \sqrt{1 - a_t^2}
\end{equation}

\noindent where $\beta_t = \beta_{min} + (\beta_{max} - \beta_{min}) \cdot t$ with $\beta_{min} = 10^{-4}$ and $\beta_{max} = 2\times 10^{-2}$. To obtain continuous time $a_t$ we employ simple linear interpolation as in \cite{zhang2023fast}. We use the AdamW optimizer with learning rate $10^{-4}$ and no gradient clipping. We also use the standard process of using \textit{Exponential Moving Average (EMA)} while training the network $\b \epsilon_{\theta}$ using \cref{eq:simple_loss_training}. We used a minibatch size of $128$ on each of $4$ GPUs, making the effective batch size $512$. We trained for $2000$ epochs on both CIFAR-10 \cite{cifar} and LSUN-Church constituting $196k$ iterations and simply chose the final checkpoint. For faster training, we used mixed-precision training, which did not degrade any performance as per our experiments. The architecture of the U-Net used as $\b \epsilon_{\theta}$ is taken exactly to be the standard architecture proposed in iDDPM \cite{iddpm}, with a dropout rate of $0.3$. All our experiments are implemented using the \texttt{diffusers} library \cite{von-platen-etal-2022-diffusers}.\footnote{\url{https://github.com/huggingface/diffusers/tree/main}}

When generating samples for evaluation, we set the random seed to be the same value across all experiments. This allows a better comparison between different ODE sampling methods as different samplers will still follow similar trajectories (see \cref{fig:generation_comparison}).

\section{Mathematical Simplifications}

We note that a number of simplifications/analytic results can be leveraged for DEIS, when applied specifically to the variance preserving process \cite{ddpm,zhang2023fast},

\begin{equation}
\sigma_t^2=1-a_t^2,\quad g_t^2 = -2f_t, \quad \Psi(t,s)=a_t/a_s.
\end{equation}

\section{DEIS-SN Implementation Details}

We follow \citet{zhang2023fast}'s DEIS implementation\footnote{\url{https://github.com/qsh-zh/deis/tree/main/th_deis}} closely for the most part. One minor difference is that we set $t_0=0$ rather than a small value such as $10^{-4}$. We find that for the samplers that we use this does not lead to any discontinuities/divide-by-zero errors, since no function is actually evaluated at $t=0$ (\cref{eq:ei} uses a one-sided Riemann sum for numerical integration).

We find empirically that performance is improved by truncating $\bar{s}_\theta(t)$ slightly, close to $t=0$. This is possibly due to numerical instability from its rapid increase as $\sigma_t\rightarrow0$. We simply set  $\bar{s}_\theta(t) = \bar{s}_\theta(0.005)$ for $t<0.005$. We calculate  $\bar{s}_\theta(t)$ by measuring the average absolute pixel values of $\b s_\theta(\b x_t,t)$ at each time step using DEIS-$t$AB3 with 1000 NFEs (1000 uniform time steps) over a batch of 256 generations. This is done using a different random seed to the generations used for evaluation. We then use linear interpolation to obtain values over continuous time.

\section{More generation results} \label{app:more_gen_results}

\begin{figure}[h]
    \centering
    \includegraphics[width=0.99\linewidth]{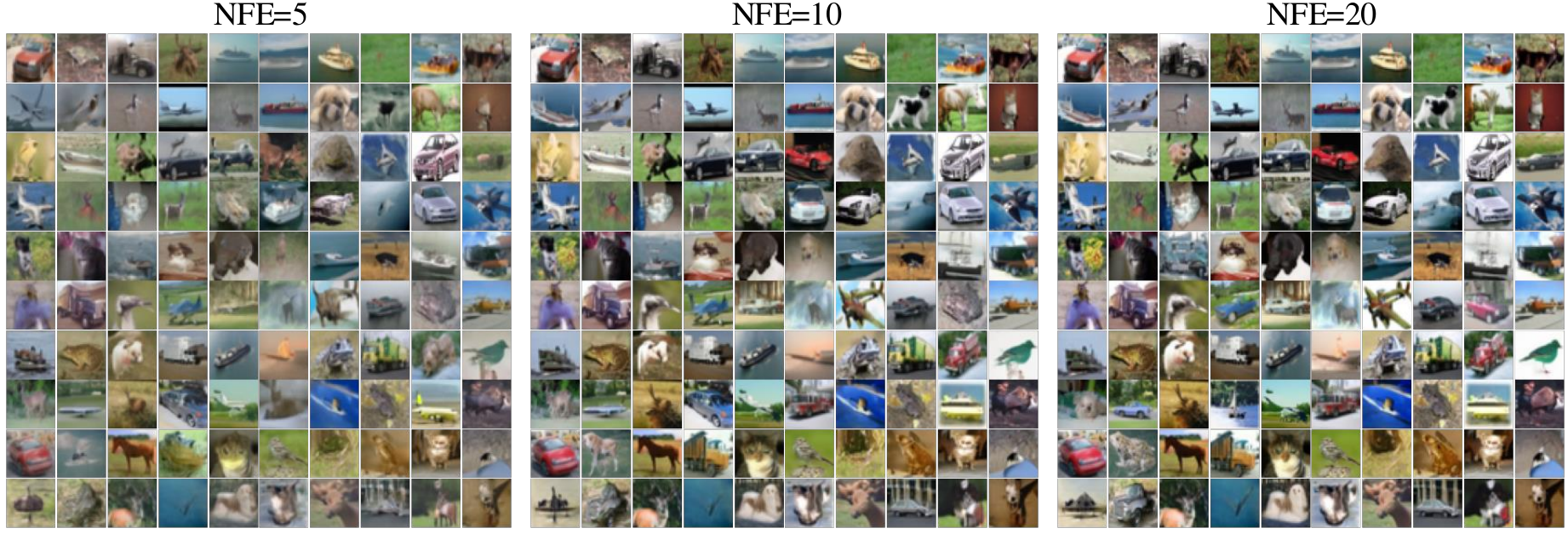}
    \includegraphics[width=0.99\linewidth]{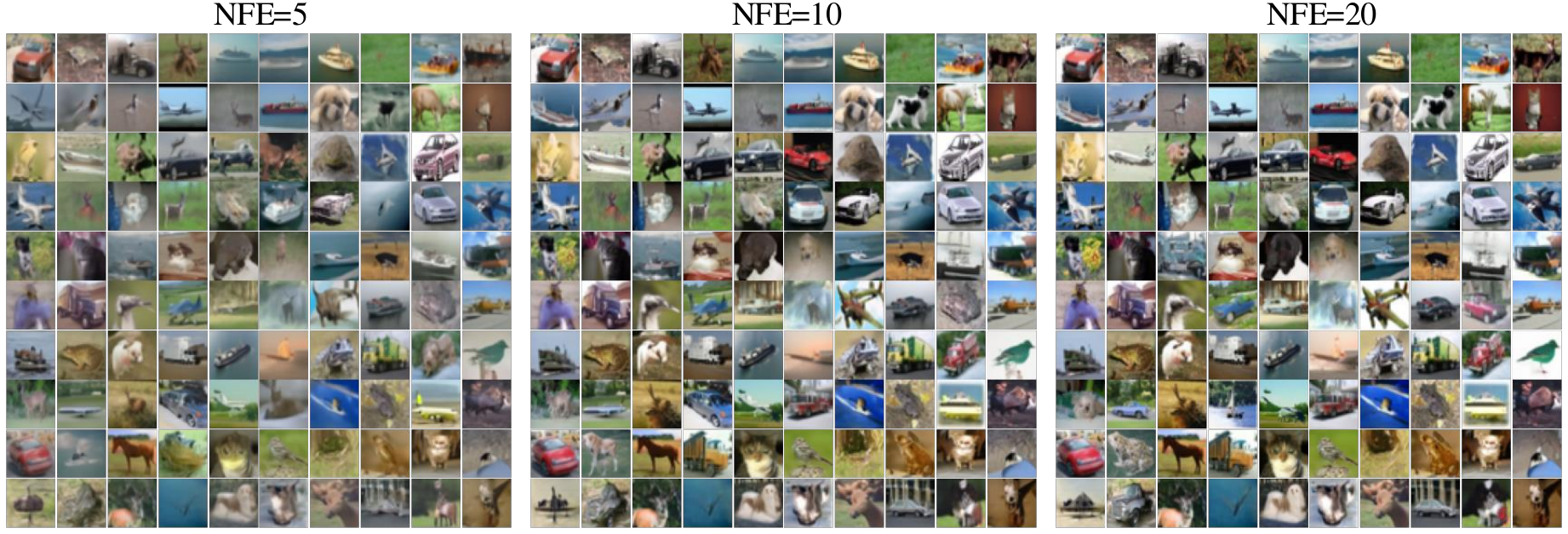}
    \includegraphics[width=0.99\linewidth]{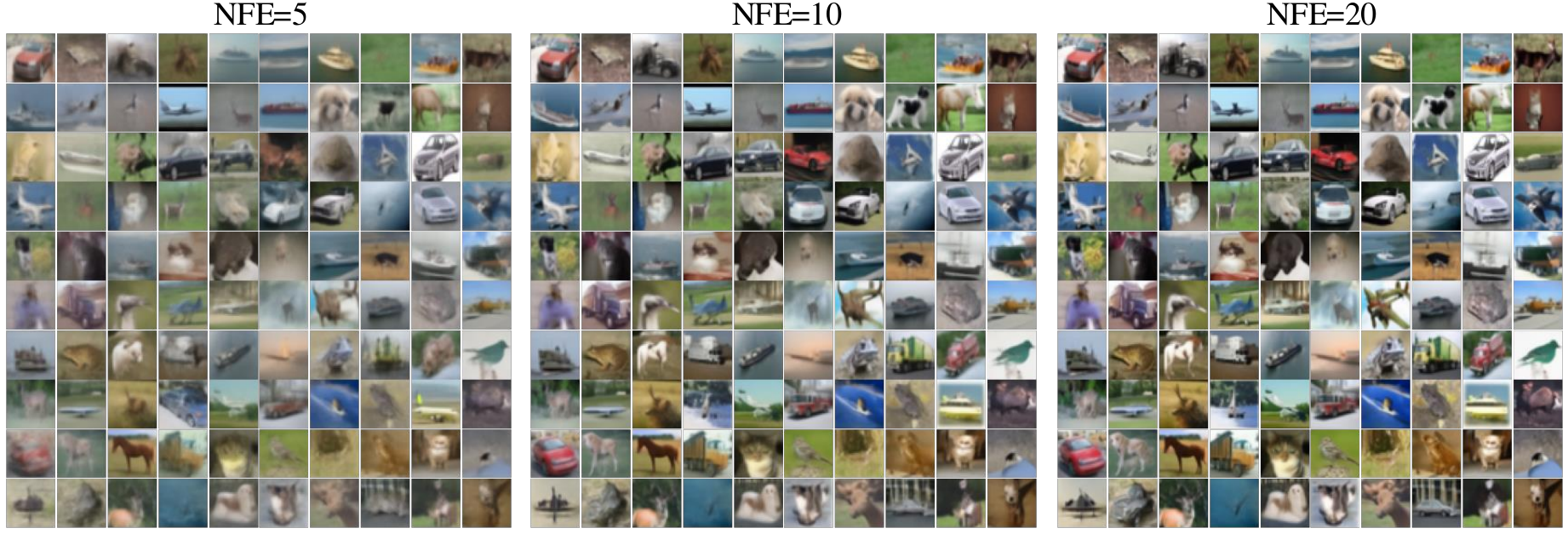}
    \includegraphics[width=0.99\linewidth]{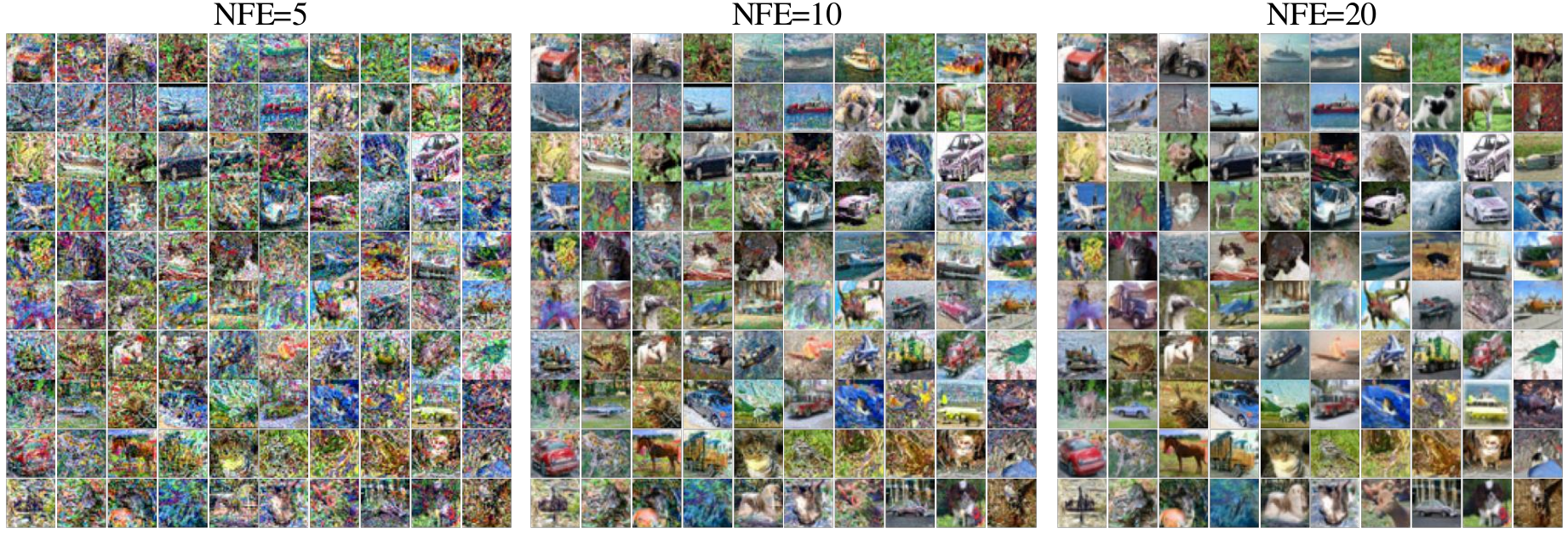}
    \caption{Visual comparison of unconditional samples for CIFAR10 generated (with same seed) at relatively low NFE for (top to bottom) \textbf{DEIS-SN (Ours)}, DEIS, DDIM \& Euler sampler.}
    \label{fig:unconditional_samples}
\end{figure}

\newpage

\section{Results on LSUN-Church} \label{sec:lsunchurch}

\begin{figure}[h]
    \centering
    \includegraphics[width=0.49\linewidth, trim={0.7cm 0 1.2cm 0},clip]{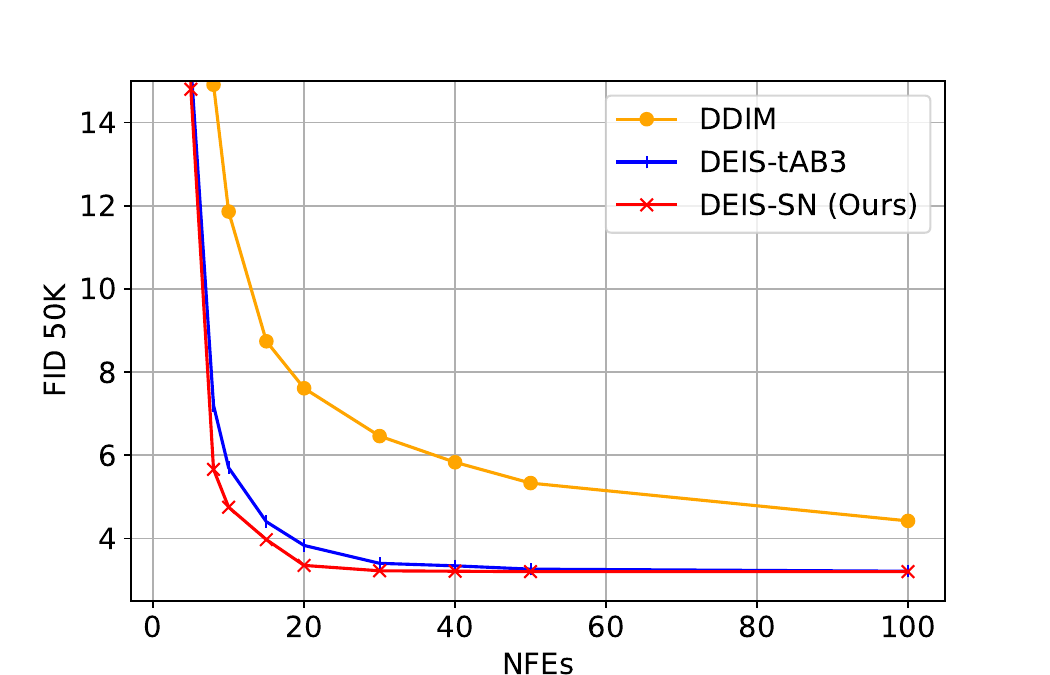}%
    \includegraphics[width=0.49\linewidth,trim={1.2cm 0 0.7cm 0},clip]{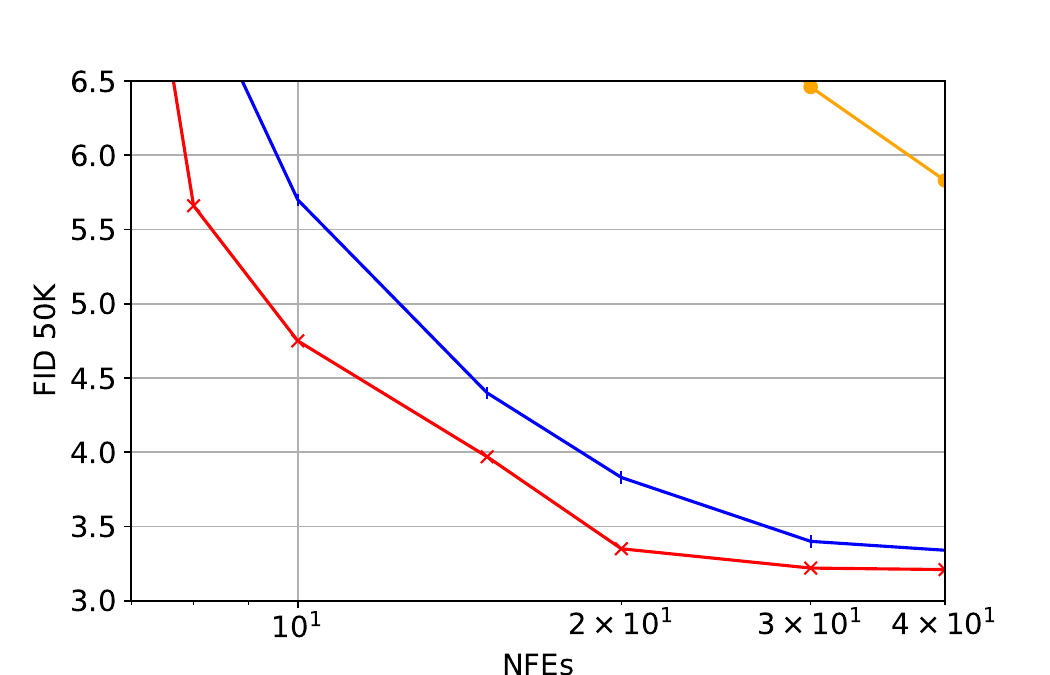}
    \caption{\textcolor{gray}{Left}: FID 50K 
 for LSUN-Church against NFEs for different samplers on the same trained model. \textcolor{gray}{Right}: zoomed in view of the left figure. \textbf{DEIS-SN consistently outperforms DEIS at low NFE regime.}}
    \label{fig:NFE-FID-Church}
\end{figure}

We follow exactly same protocol except the UNet architecture, which in this case is adapted from \cite{dhariwal2021diffusion}. The UNet in question has an attention resolution of only $16$ (unlike $16,8$ in CIFAR-10 model) and $2$ ResNet blocks (unlike $3$ in CIFAR-10 model). We also use a dropout of $0.1$ while training out LSUN-Church model. Below, in \cref{fig:NFE-FID-Church}, we present the FID-vs-NFE curves similar to \cref{fig:generation_comparison} in the main paper.

\end{document}